\documentclass[letterpaper, 10 pt, conference]{ieeeconf} 
\IEEEoverridecommandlockouts
\def\spacesavermode{2}
\AtBeginDocument{%
\ifcase\spacesavermode
\or
  \setlength{\textfloatsep}{14pt plus 4pt minus 4pt}   
  \setlength{\floatsep}{8pt plus 3pt minus 3pt}        
  \setlength{\intextsep}{8pt plus 3pt minus 3pt}       
  \setlength{\abovecaptionskip}{4pt}                   
  \setlength{\belowcaptionskip}{0pt}                   
  \setlength{\abovedisplayskip}{5pt plus 3pt minus 2pt}     
  \setlength{\belowdisplayskip}{5pt plus 3pt minus 2pt}
  \setlength{\abovedisplayshortskip}{2pt plus 3pt minus 1pt}
  \setlength{\belowdisplayshortskip}{4pt plus 3pt minus 2pt}
\or
  \setlength{\textfloatsep}{8pt plus 2pt minus 2pt}
  \setlength{\floatsep}{5pt plus 2pt minus 2pt}
  \setlength{\intextsep}{7pt plus 2pt minus 2pt}
  \setlength{\abovecaptionskip}{3.5pt}
  \setlength{\belowcaptionskip}{0pt}
  \setlength{\abovedisplayskip}{6pt plus 2pt minus 2pt}
  \setlength{\belowdisplayskip}{6pt plus 2pt minus 2pt}
  \setlength{\abovedisplayshortskip}{1pt plus 1pt minus 1pt}
  \setlength{\belowdisplayshortskip}{2pt plus 1pt minus 1pt}
\or
  \setlength{\textfloatsep}{6pt plus 2pt minus 2pt}
  \setlength{\floatsep}{4pt plus 2pt minus 2pt}
  \setlength{\intextsep}{6pt plus 2pt minus 2pt}
  \setlength{\abovecaptionskip}{3pt}
  \setlength{\belowcaptionskip}{0pt}
  \setlength{\abovedisplayskip}{4pt plus 2pt minus 2pt}
  \setlength{\belowdisplayskip}{4pt plus 2pt minus 2pt}
  \setlength{\abovedisplayshortskip}{2pt plus 2pt minus 1pt}
  \setlength{\belowdisplayshortskip}{3pt plus 2pt minus 1pt}
\fi
}%
\usepackage{amssymb}
\usepackage{amsmath}
\usepackage{algorithm}
\usepackage{algpseudocode}
\usepackage[caption=false,font=footnotesize]{subfig}
\usepackage{graphicx}
\newcommand{\hdr}[2]{\shortstack[c]{#1\\(\S\ref{#2})}}

\title{\LARGE \bf
High-Altitude Balloon Station-Keeping with
First Order Model Predictive Control
}

\author{Myles Pasetsky$^{1*}$, Jiawei Lin$^{2*}$, Bradley Guo$^{1*}$, and Sarah Dean$^{1}$%
\thanks{$^{1}$Cornell University
        {\tt\small \{mhp58, bzg4, sdean\}@cornell.edu}}%
\thanks{$^{2}$University of California San Diego {\tt\small jal214@ucsd.edu}}
\thanks{$^{*}$Equal Contribution}%
}

\usepackage{xcolor}
\usepackage{float}
\usepackage{hyperref}
\usepackage{caption}
\begin{document}

\maketitle
\thispagestyle{empty}
\pagestyle{empty}

\begin{abstract}

High-altitude balloons (HABs) are common in scientific research due to their wide range of applications and low cost. Because of their nonlinear, underactuated dynamics and the partial observability of wind fields, prior work has largely relied on model-free reinforcement learning (RL) methods to design near-optimal control schemes for station-keeping. These methods often compare only against hand-crafted heuristics, dismissing model-based approaches as impractical given the system complexity and uncertain wind forecasts. We revisit this assumption about the efficacy of model-based control for station-keeping by developing \textit{First-Order Model Predictive Control (FOMPC)}. By implementing the wind and balloon dynamics as differentiable functions in JAX, we enable gradient-based trajectory optimization for online planning. FOMPC outperforms a state-of-the-art RL policy,
achieving a 24\% improvement in time-within-radius (TWR) without requiring offline training, though at the cost of greater online computation per control step. Through systematic ablations of modeling assumptions and control factors, we show that online planning is effective across many configurations, including under simplified wind and dynamics models.

\end{abstract}

\section{INTRODUCTION}
High-altitude balloons (HABs) provide a low-cost and low-risk platform \cite{picouet2025stratospheric} for atmospheric and climate research \cite{habeck2020high,sushko2017low}, as well as astronomical and space science observations \cite{ross1958strato,klomchitcharoen2024high}, among others. As horizontal control remains costly and resource-intensive, most of these balloons rely on active altitude control to take advantage of varying wind currents at different altitudes that guide the balloons laterally. However, accurate measurements of these wind currents are only available locally, making the setting partially observable. In this underactuated and partially observable setting, control effectiveness depends both on the accuracy of wind predictions and the reliability of altitude-based steering. These requirements are further complicated by internal resource limitations, which can restrict altitude control, and by external environmental factors, which may reduce the availability of diverse and favorable wind layers.

Early work on developing controllers for autonomous HABs relied on hand-designed heuristics~\cite{du2019energy, du2019station}, which explicitly exploit the local wind column via altitude control to achieve short-term station-keeping.  
To approach the policy from an optimal control perspective, recent works apply model-free reinforcement learning (RL) methods to the balloon control problem \cite{jeger2023reinforcement,xu2022station,he2025path,saunders2023resource,gannetti2023navigation,liu2024high,bellemare2020autonomous, garg2019wind}. These approaches train policies in simulation to either navigate from point to point in a wind field or perform station-keeping, where the balloon remains in proximity to a ground target. Among these works, the RL agent has access only to the local predicted wind profile (typically a vertical column) at its current horizontal position and learns an approximate optimal policy through offline training in simulation.

Despite successful RL deployments in both simulation and on hardware, their main performance gains are demonstrated only with respect to simple heuristic-based policies. These works are often motivated by the purported difficulty of applying model-based tools to the station-keeping problem \cite{bellemare2020autonomous}, 
and the need for costly replanning under uncertain wind predictions \cite{jeger2023reinforcement}.
Moreover, strong baselines are critical for validating empirical results in deep RL, where reported gains can be undermined by statistical uncertainty with few evaluation runs~\cite{agarwal2021deep} and high training variance across seeds~\cite{mania2018simple}. Without strong baselines, deep RL results remain difficult to interpret, as costly training, variance across seeds, and generalization failures can exaggerate apparent progress.
Hence, there is a critical gap in the literature: a systematic model-based baseline for direct comparison with RL methods. 

\begin{figure}[t]
    \centering
    \includegraphics[width=\columnwidth]{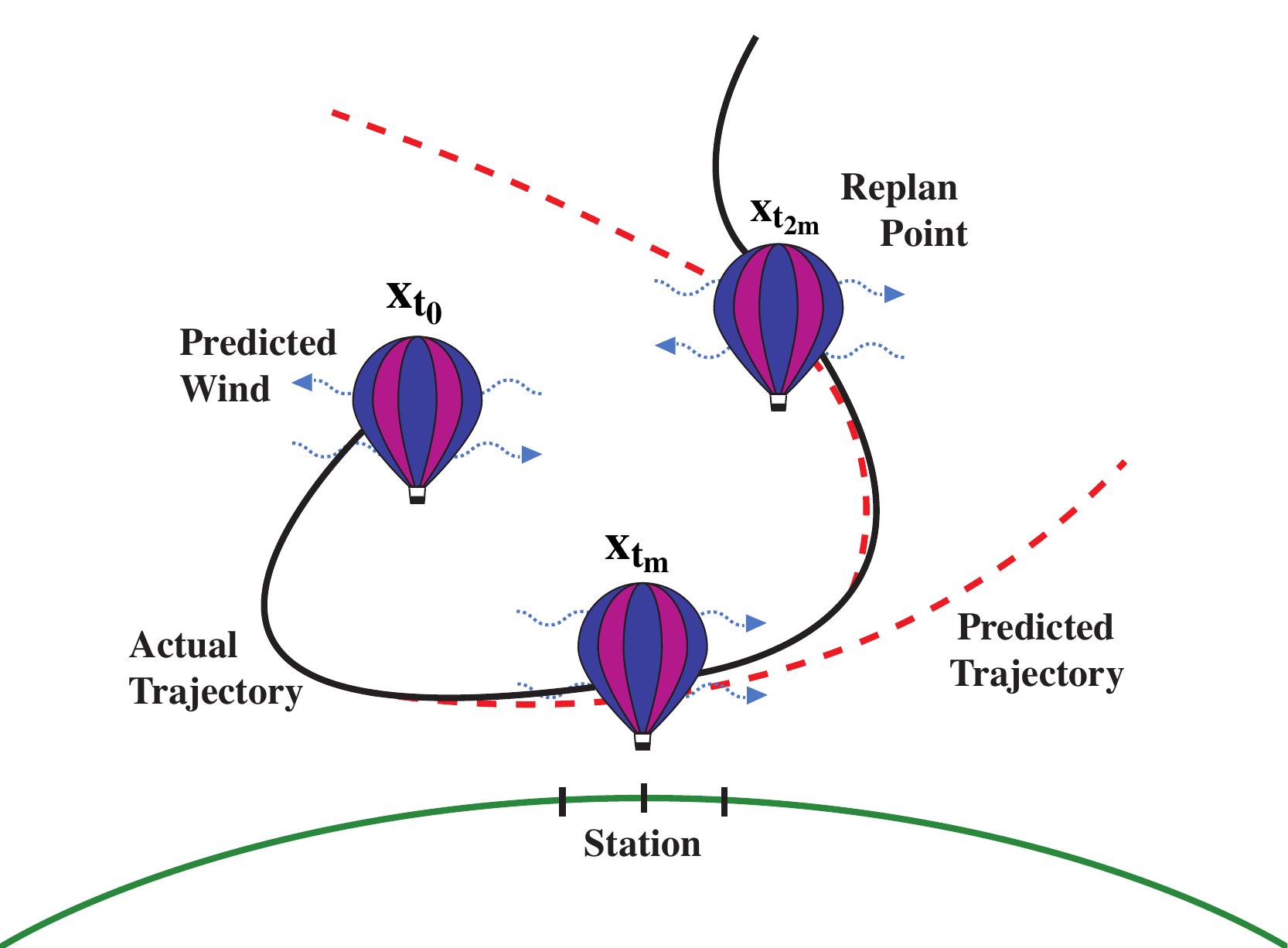}
    \caption{Station-keeping via receding-horizon altitude control. Predicted winds generate a planned trajectory (red), which diverges from the actual trajectory (black); replanning at every $m$ steps corrects forecast errors.}
    \label{fig:your_label}
\end{figure}

Motivated by this absence of model-based baselines, we develop First-Order Model Predictive Control (FOMPC) as the first rigorous approach to fill this gap. By first implementing balloon dynamics and wind models as differentiable functions in JAX~\cite{jax2018github}, then formulating the HAB station-keeping task as an unconstrained optimization problem, we can use first-order gradient methods to optimize balloon actions. This approach explicitly incorporates both dynamics and wind predictions into long planning, enabling the controller to exploit future wind patterns rather than reacting only to current wind conditions. Frequent replanning then mitigates forecast errors from imperfect wind predictions and dynamics fidelity, allowing even simplified dynamics and wind models to outperform model-free RL policies. 

We evaluate FOMPC in the Balloon Learning Environment (BLE)~\cite{Greaves_Balloon_Learning_Environment_2021} against Google's state-of-the-art RL controller, Perciatelli44~\cite{agarwal2022reincarnating}, and show that FOMPC achieves up to a 24\% improvement in time-within-radius (TWR), or 1.8 extra hours a day in radius, with the same wind information. Through systematic ablations on planning horizon, replan interval, model fidelity, and wind models, we further identify the factors most responsible for performance gains and robustness. Our main contributions are:

\begin{itemize}
    \item We release an open-source, fully differentiable implementation of high-altitude balloon dynamics in JAX\footnote{\href{https://github.com/sdean-group/balloon-learning-environment}{https://github.com/sdean-group/balloon-learning-environment}} to support future research in model-based control and learning.
    \item We propose First-Order Model Predictive Control (FOMPC), which outperforms Perciatelli44 by 24\% in time-within-radius performance, an additional 1.8 hours a day within radius,  and establishes a systematic model-based baseline for station-keeping.
    \item We conduct a comprehensive ablation study of modeling and control parameters, identifying which factors most strongly drive performance and where simplified models remain effective. These insights provide concrete guidance for future RL and hybrid approaches.
\end{itemize}

\section{RELATED WORK}
Most prior research on HAB station-keeping focuses on model-free RL, motivated by the perceived difficulty of applying model-based approaches under uncertain winds. Non-RL alternatives remain limited, typically relying on heuristic-based or data-driven methods, and rarely compared against RL. Meanwhile, model predictive control (MPC) in broader aerospace applications shows that MPC can perform well under wind disturbances and nonlinear dynamics, suggesting it to be a promising but underexplored baseline for the HAB station-keeping problem. 

\subsection{Reinforcement Learning Approaches}
Most prior work focuses on reinforcement learning (RL) methods. Project Loon addressed the station-keeping problem using quantile regression deep Q-network (QR-DQN) to maintain balloons at altitudes around 20 km for long durations while providing Internet connectivity \cite{bellemare2020autonomous}. Later, they further improved efficiency by reusing prior computation and experience to accelerate learning, demonstrated in their Balloon Learning Environment \cite{agarwal2022reincarnating}. Subsequent work applied variants of DQN~\cite{xu2022station,gannetti2023navigation,liu2024high} and SAC~\cite{saunders2023resource} to tasks including station-keeping, navigation, and no-fly zone avoidance, as well as to different balloon types such as sounding and latex balloons. Collectively, these efforts show that RL can exploit wind structure without explicit models, but do not amend the view that model-based approaches are impractical under stratospheric uncertainty.

\subsection{Non-RL Alternatives}
Non-RL alternatives remain sparse. Heuristic and model-based altitude controllers have been proposed without long-horizon planning~\cite{du2019station}, and extremum-seeking controllers learn favorable winds without explicit models~\cite{harry2025autonomous}, but neither is compared to RL. While many works argue that model-based methods are fragile under underactuation and forecast errors, studies in related flow settings show that replanning can enable robust performance~\cite{wiggert2022navigating}. Most closely related, a receding-horizon controller using short-term forecasts and optimistic optimization has been demonstrated~\cite{fan2024station}, but the method relies on gradient-free optimization and is only compared against tree search, not RL.

\subsection{Model Predictive Control}

Model Predictive Control (MPC) has been widely applied in unmanned aerial systems (UAS), demonstrating robustness under wind disturbances and nonlinear dynamics. For example, constrained MPC solved via quadratic programming has stabilized quadrotor attitude and position~\cite{aliyari2022design}, while efficient nonlinear formulations have improved real-time performance by shortening prediction horizons~\cite{gomaa2022computationally}. Adaptive nonlinear MPC has been used for highly unsteady aircraft dynamics~\cite{garcia2014robust}, and cooperative MPC has enabled coordinated aerial–ground rendezvous with reduced computation~\cite{persson2017cooperative}. In lighter-than-air systems, MPC has guided stratospheric airships using linearized dynamics~\cite{zhang2019trajectory} and extended to constrained trajectory tracking under unknown disturbances with Laguerre function parameterizations~\cite{yuan2020trajectory}. These successes highlight MPC's versatility but leave open whether it can plan effectively in uncertain wind fields with the limited control authority of HABs.
\section{PROBLEM FORMULATION}
\label{sec:problem_formulation}
The goal of station-keeping is to maintain the balloon's position within a communication range of a ground station. The evaluation metric is time-within-radius (TWR), defined as the fraction of time steps during which the balloon remains within 50 km of the ground station center. We frame station-keeping as a finite-horizon discrete-time optimal control problem.
Over an episode of $K$ steps, the objective is to compute the control policy $\pi$ that maximizes the expected TWR subject to the balloon dynamics:
\begin{equation}
\begin{aligned}
\max_{\pi} \quad & \mathbb{E}\left[ \frac{1}{K}\sum_{k=0}^{K-1} 
\mathbf{1}\!\left(x_k^2 + y_k^2 \leq 50^2\right) \right]\\
\text{s.t.} \quad & \mathbf{x}_{k+1} = F_{\Delta t}(\mathbf{x}_k, \mathbf{w}_k, u_k, t_k), \\
\quad &k=0,\dots,K-1, ~~
 \mathbf{x}_0 =\mathbf{x}_{\text{init}},\\
& u_k = \pi(\mathbf x_k, \hat{W}),
\label{problem_def} 
\end{aligned}
\end{equation}
where $\mathbf{1}(\cdot)$ is the indicator of being within the 50 km radius. Here, $\mathbf{x}_k$ is the balloon state (with its horizontal components $x_k$ and $y_k$ expressed here in kilometers), $u_k \in [-1,1]$ is the vertical control input, and $t_k$ is the timestamp. $\hat{W}$ denotes the wind information available to the agent (Sec~\ref{subsec:wind_model}), while $\mathbf{w}_k$ is the wind at the balloon's position (Sec~\ref{sec:wind}). $F_{\Delta t}$ is the state update \eqref{eq:discrete_map}. The expectation is over forecast uncertainty and stochastic wind realizations. This objective is consistent with prior RL evaluations, enabling direct comparison.

\section{SYSTEM DYNAMICS MODEL}
We describe balloon dynamics as described in BLE \cite{Greaves_Balloon_Learning_Environment_2021}. These are re-implemented in JAX with an additional continuous control relaxation for differentiability. HAB dynamics can be decomposed into horizontal dynamics, governed solely by the wind, and altitude dynamics, determined by buoyancy and net mass.

Let $\mathbf{x}$ define the state vector of the superpressure balloon. 
{\setlength{\abovedisplayskip}{4pt}
 \setlength{\belowdisplayskip}{4pt}
\begin{equation}
\mathbf{x}=[x, y, l, n, T, E, V, p_{\text{env}}],
\end{equation}}
where $(x,y)$ are horizontal offsets (stored in meters), $l$ pressure, $n$ ballonet air moles, $T$ temperature, $E$ battery energy, $V$ envelope volume, and $p_{\text{env}}$ envelope superpressure. 

Dynamics follow
{\setlength{\abovedisplayskip}{4pt}
 \setlength{\belowdisplayskip}{4pt}
\begin{equation}
\dot{\mathbf{x}}=f(\mathbf{x},\mathbf{w},u,t),
\end{equation}}
implemented with discrete updates computed with $\Delta t=N\delta t$ via $N$ Euler substeps ($\delta t=10$s, $N=18$) as follows:

{\setlength{\abovedisplayskip}{4pt}
 \setlength{\belowdisplayskip}{4pt}
 \begin{equation}
\begin{aligned}
\mathbf{x}^{(j+1)} &= \mathbf{x}^{(j)} + \delta t
      f(\mathbf{x}^{(j)}, \mathbf{w}_k, u_k, t_k+j\delta t) \\
    j&=0, \cdots, N-1, ~~\mathbf{x}^{(0)}=\mathbf{x}_k
\end{aligned}
\end{equation}}
The final state after one simulator step is $\mathbf{x}_{k+1}=\mathbf{x}^{(N)}$. This composition defines our discrete update step:
{\setlength{\abovedisplayskip}{4pt}
 \setlength{\belowdisplayskip}{4pt}
\begin{equation}\label{eq:discrete_map}
    \mathbf{x}_{k+1}=F_{\Delta t}(\mathbf{x}_k,\mathbf{w}_k,u_k,t_k)
\end{equation}}

\subsection{Wind Fields in Environment}\label{sec:wind}
BLE generates forecast winds $W_0$ from a variational autoencoder (VAE) trained on the ERA5 reanalysis dataset~\cite{hersbach2017era5}. The ground truth wind, $W$, is defined with added Simplex noise $b$ to emulate forecast errors:
\begin{equation}
\label{eq:wind_field_def}
W(x,y,l,t) = W_0(x,y,l,t) + b(x,y,l,t).
\end{equation}

We also define $\mathbf{w}_k\equiv W(x_k,y_k,l_k,t_k)$.

\subsection{Balloon Dynamics} 
We decompose balloon dynamics into six components, consistent with the Balloon Learning Environment (BLE) simulator. 
\subsubsection{Horizontal Dynamics}
The balloon drifts with the ambient wind at its current location and pressure level. 
\[
\dot{x} = \mathbf{w}_x, \quad\dot{y}=\mathbf{w}_y
\]

\subsubsection{Vertical Dynamics}
Let $h$ denote the altitude of the balloon. Vertical motion is defined by buoyant, gravity, and drag forces:
\begin{equation}
m\ddot{h}=F_b-F_g-F_d,
\end{equation}
but BLE enforces steady state ($\ddot{h}=0$):
\begin{equation}
0=\rho_{\text{air}} Vg - mg - \tfrac{1}{2}\rho_{\text{air}} C_dA|\dot{h}|\dot{h}\label{eq:vertical_balance}.
\end{equation}

We can directly solve for $\dot{h}$ from \eqref{eq:vertical_balance}. Using the U.S. Standard Atmosphere Model~\cite{united1976us}, we convert $\dot{h}$ into a pressure differential, $\dot{l}$, which allows us to fully model pressure dynamics.
    
\subsubsection{Thermal Model}\label{sec:thermal}
The envelope temperature $T$ evolves from ambient temperature, solar flux, and radiation balance, which determines the density of air inside the envelope. Warmer internal temperature reduces gas density, increases buoyancy, and thus promotes ascent; vice versa. 

\subsubsection{Volume and Superpressure Model}\label{sec:vol}
In \eqref{eq:vertical_balance}, a larger volume $V$ directly increases buoyancy and promotes ascent. Superpressure $p_{\text{env}}$ limits control: once fully stretched, added gas/heat raises $p_{\text{env}}$ not $V$, increasing venting when high and risking collapse when low.

\subsubsection{Altitude Control System (ACS)}\label{sec:acs}
The ACS adjusts $n$, the moles of air, via a continuous action $u\in[-1,1]$, with venting for $u>0$ and pumping for $u<0$:
\begin{equation}
\dot{n}(u)=
\begin{cases}
-\,c_{\text{vent}}(l,T,p_{\text{env}})u, & u > 0,\\
c_{\text{pump}}(l,T,p_{\text{env}})(-u), & u < 0,\\
0, & u=0,
\end{cases}
\end{equation}
with power
\begin{equation}
P_{\text{acs}}(u)=
\begin{cases}
\eta(l,p_{\text{env}})(-u), & u < 0,\\
0, & u \geq 0.
\end{cases}
\end{equation} 

This is a relaxation compared with the original discrete actions defined in BLE.
It allows our algorithm to plan a continuous control input $u$, which is mapped into physically realistic venting or pumping rates.\footnote{Re-discretizing post-optimization yields similar time-within-radius performance, supporting the validity of the continuous relaxation. Using the $N$ Euler substeps to approximate continuous actions as multiples of $\frac{1}{N}$ by alternating discrete actions also produced comparable results.}

\subsubsection{Battery Dynamics}\label{sec:battery}
Battery energy evolves as $\dot{E}=P_{\text{solar}}(t)-P_{\text{load}}(t)-P_{\text{acs}}(u)$, with BLE disabling downward pumping when $E/E_{\max}<0.025$ until recovery above $0.05$ (considered a power safety violation).

\begin{figure*}[t] 
    \centering
    \includegraphics[width=\textwidth]{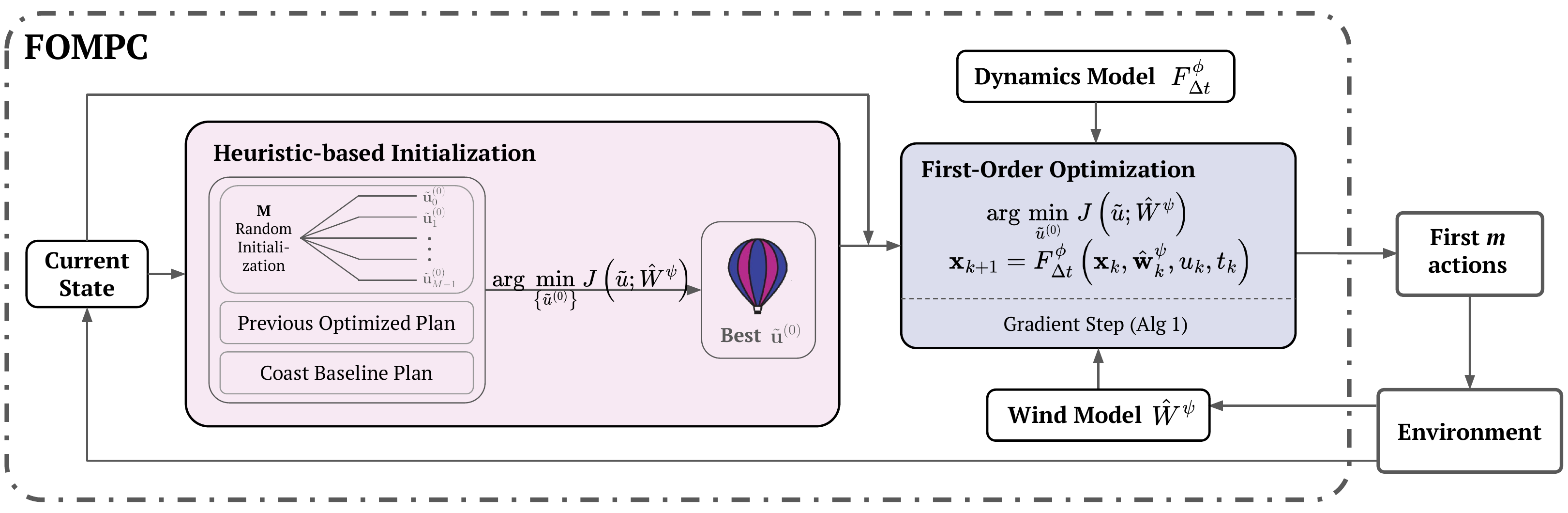}
    \caption{Block diagram of FOMPC. Candidate plans are initialized using heuristics, then refined by first-order optimization with dynamics model $F_{\Delta t}^\phi$ and wind model $\hat{W}^\psi$. The lowest-cost plan provides the first $m$ actions, after which the process repeats with the updated state.}
    \label{fig:wide}
\end{figure*}

\section{MODEL PREDICTIVE CONTROL DESIGN}
\let\savedthesubsubsection\thesubsubsection
\renewcommand{\thesubsubsection}{\thesection.\arabic{subsubsection}}

MPC is a control strategy that optimizes actions over a receding horizon. Typical MPC formulations attempt to simplify the planning problem into a convex surrogate. However, iterative approximations and standard nonlinear optimization techniques can be computationally prohibitive~\cite{zhang2024-dr}. Instead, we reformulate the planning task as an unconstrained optimization problem. This allows us to leverage gradient-based optimization to solve the problem with significantly higher efficiency.
We  propose \emph{First-Order MPC (FOMPC)}, which we describe by its main components:

\subsubsection{Continuous Action Construction}
\label{subsec:cont_action_constriction}
Since control actions are only defined for $u \in [-1, 1]$, we parameterize them as $u = \varsigma(\tilde u) = \tanh(\tilde u/2)$. Thus, optimization is over unconstrained $\tilde{u}$, while $u$ remains admissible by construction.

\subsubsection{Dynamics Model}
In FOMPC, the planning dynamics $F_{\Delta t}^\phi$ need not match the full BLE dynamics $F_{\Delta t}$. We define reduced-order models by omitting selected physical states and updates. Each fidelity level $\phi\in\{\phi_0,\ldots,\phi_4\}$ specifies which processes are integrated for each $\delta t$ and which are clamped (Table \ref{tab:fidelity}). The resulting dynamics are

\begin{equation}
    \mathbf{x}_{k+1} = F^{\phi}_{\Delta t}(\mathbf{x}_k, \mathbf{w}_k, u_k, t_k),
\end{equation}
where $F^{\phi}_{\Delta t}$ applies $N$ sub-steps of Euler integration with the subset of updates determined by $\phi$.
\begin{table}[H]
\centering
\setlength{\tabcolsep}{6pt}
\renewcommand{\arraystretch}{1.15}
\begin{tabular}{c|c c c c}
Level & \hdr{ACS}{sec:acs} & \hdr{Vol/Superpres.}{sec:vol} & \hdr{Temp.}{sec:thermal} & \hdr{Batt.}{sec:battery} \\
\hline
$\phi_0$ & $\times$ & $\times$ & $\times$ & $\times$ \\
$\phi_1$ & $\checkmark$ & $\times$ & $\times$ & $\times$ \\
$\phi_2$ & $\checkmark$ & $\checkmark$ & $\times$ & $\times$ \\
$\phi_3$ & $\checkmark$ & $\checkmark$ & $\checkmark$ & $\times$ \\
$\phi_4$ & $\checkmark$ & $\checkmark$ & $\checkmark$ & $\checkmark$
\end{tabular}
\caption{Fidelity levels and enabled submodules}
\label{tab:fidelity}
\end{table}
When clamped, the values of states in a submodule are held to their values in $\mathbf{x_0}$. As the ACS is not a state variable directly, we instead replace it with a piecewise-linear surrogate 

\begin{equation}
\dot{n}(u)=
\begin{cases}
-0.012u, & u > 0,\\
0.007(-u), & u < 0,\\
0, & u=0,
\end{cases}
\end{equation}
and linearize the power usage similarly, $P_{\text{acs}}(u) = 195(-u)$ for $u<0$ (with usage remaining $0$ for $u \ge 0$).
These constants are averages derived from typical ACS operating conditions in BLE.

\subsubsection{Wind Model}
\label{subsec:wind_model}
The balloon's horizontal dynamics depend on the ground-truth wind $W$, while MPC plans, at best, can have an approximate model $\hat{W}^\psi$. Because the forecast $W_0$ contains systematic errors, we construct several wind models for investigation.

We define four variants:
\begin{enumerate}
    \renewcommand{\theenumi}{(\roman{enumi})}
    \renewcommand{\labelenumi}{\theenumi}
    \item $\hat{W}^{\psi_0} = W_0$, where MPC can plan through the forecast.
    \item $\hat{W}^{\psi_1}_{\mathbf{x}_0}(x,y,l, t) = W_0(x_0, y_0, l, t_0)$, where $W_0$ is  treated as constant across $(x,y,t)$, forming a wind column centered around the balloon state at the start of a replan.
    \item $\hat{W}^{\psi_2}_{\mathbf{x}_0}(x,y,l, t) = W_0(x_0,y_0,l,t_0) +\mathcal{G}(x_0, y_0, l, t_0)$, where $\mathcal{G}$ is a Gaussian Process trained to predict forecast noise from forecast errors $\{ \mathbf{{w}}_{0:t} - \mathbf{w_0}_{0:t} \}$ sampled at the locations $\{ \mathbf{x}_{0:t} \}$ traversed since previous replan, from ~\cite{bellemare2020autonomous}.
    \item $\hat{W}^{\psi_3}_{\mathbf{x}_0} = \alpha \cdot W_0 + (1-\alpha)\cdot \hat{W}^{\psi_2}_{\mathbf{x}_0}$, which is a combination of the forecast and the processed wind column to take advantage of online re-estimation of winds while still incorporating the wind variation across $(x,y,t)$. We set $\alpha=0.5$.
\end{enumerate}


As described in \cite{bellemare2020autonomous} and implemented in \cite{Greaves_Balloon_Learning_Environment_2021}, the GP uses a Matérn kernel and is trained once per replan at $\mathbf{x}_0$, returning estimated wind forecast errors. When combined with the original forecast, this provides an online estimate of the local wind forecast we use in $\hat{W}^{\psi_2}$ and $\hat{W}^{\psi_3}$.

\subsubsection{Differentiable Cost Function Design}
To replace the non-smooth TWR objective, we use a differentiable surrogate penalizing the squared distance and low battery:
\begin{equation}
    c(\mathbf{x}) = \left(\tfrac{x}{1000}\right)^{2} + \left(\tfrac{y}{1000}\right)^{2} \;+\; R^{2}c_{\text{power}}(\mathbf{x})
\label{eq:cost_fn}
\end{equation}
where $c_{\text{power}}(\mathbf{x}) = 1 - \sigma\!\left(100 \left(\tfrac{E}{E_{\text{max}}} - 0.1\right)\right)$, $\sigma(z)=1/(1+\exp(-z))$, and $R=50\text{km}$ matches the TWR radius. The $c_{\text{power}}$ term begins penalizing usage around 10\% capacity, approximating the power safety layer that activates at 2.5\% and 5\% (Sec.~\ref{sec:battery}).

\subsubsection{Receding-horizon Planning}

At replanning instant $t_n$, with wind model $\hat{W}^\psi$, dynamics model $F^\phi_{\Delta t}$, current balloon state $\mathbf{x}(t_n) $, planning horizon $H$ and discount factor $\gamma$, we define the optimization objective Eq.~\ref{eq:mpc}.

{\setlength{\abovedisplayskip}{6pt plus 2pt minus 2pt}
\setlength{\belowdisplayskip}{6pt plus 2pt minus 2pt}
\begin{table}

\begin{equation}
\begin{aligned}
\min_{\{\tilde {u}_k\}} \quad & J_n(\{\tilde{u}_k\},  \mathbf{x}_0;\hat{W}^\psi) := \sum_{k=0}^{H-1} \gamma^{k}\, c(\mathbf{x}_{k+1}) \label{eq:mpc}\\
\text{s.t.}\quad & \mathbf{x}_{k+1}=F^\phi_{\Delta t}(\mathbf{x}_k,\hat{\mathbf{w}}_k^\psi, u_k,t_k), \quad k=0,\dots,H-1 \\
& u_k=\varsigma(\tilde {u}_k)\,\mathbf{1}\!\left(\frac{E_k}{E_\text{max}} \ge 0.025\right), \quad \mathbf{x}_0=\mathbf{x}(t_n)
\end{aligned}
\end{equation}
\end{table}}

Substituting dynamics and control updates yields an unconstrained direct shooting formulation. We approximately solve for $H$ actions, executing the first $m$ before the next replanning cycle.


\subsubsection{Heuristic-based Plan Initialization}

To mitigate nonconvexity, FOMPC selects its starting point from $M$ random candidates, the previous optimized plan, and a ``coast'' baseline drawn from $\mathrm{Unif}(-0.2, 0.2)$
 noise, which keeps the balloon nearly stationary while remaining differentiable. Random candidates are altitude-seeking maneuvers. We first simulate a full ascent and a full descent from the current state and record the number of steps $\hat{t}(h)$ at which each altitude
$h$ is reached, up to the altitude bounds of $[15.4,19.1]$ km (defined in BLE~\cite{Greaves_Balloon_Learning_Environment_2021}). Each candidate then samples a target altitude uniformly within these bounds, applies near-saturated controls for the interpolated $\hat{t}(h)$ steps, and fills the remainder of the horizon with small random perturbations. The lowest-cost candidate initializes gradient descent.

\subsubsection{Gradient Descent Optimization}
We implement the first-order optimization directly on unconstrained $\tilde{u}$.
\begin{algorithm}[H]
\caption{First-order gradient descent}
\label{alg:fogd}
\begin{algorithmic}[1]
\State \textbf{Input:} horizon $H$, step size $\eta>0$, tolerance $\epsilon>0$, max iters $S_{\max}$, wind model $\hat{W}^\psi$, heuristic-based initialization $\tilde{\mathbf{u}}^{(0)}$
\For{$s = 0$ to $S_{\max}$}
    \State $g^{(s)} \gets \nabla_{\tilde{\mathbf{u}}} \, J_n(\mathbf{\tilde{u}}, \mathbf{x}_0; \hat{W}^\psi)\big|_{\tilde{\mathbf{u}}=\tilde{\mathbf{u}}^{(s)}}$
    \If{$\|g^{(s)}\|_2 < \epsilon$}
        \State \textbf{break}
    \EndIf
    \State $\tilde{\mathbf{u}}^{(s+1)} \gets \tilde{\mathbf{u}}^{(s)} - \eta \, \dfrac{g^{(s)}}{\|g^{(s)}\|_2}$
\EndFor
\State \textbf{Output:} $\{u_k\}_{k=0}^{H-1} \gets \{\varsigma(\tilde{u}_k^{(s)})\}_{k=0}^{H-1}$
\end{algorithmic}
\end{algorithm}

For notation simplicity, we define $\tilde{\mathbf{u}}^{(s)} :=\{\tilde{u}_k^{(s)}\}_{k=0}^{H-1}$. We solve~\eqref{eq:mpc} with gradient descent, computing $\nabla_{\tilde u}J$ by autodiff (JAX), demonstrated in Algorithm~\ref{alg:fogd}. We use step size $\eta=1$, $S_{\max}=100$, tolerance $\epsilon=10^{-7}$.
\let\thesubsubsection\savedthesubsubsection

\section{EXPERIMENTS}

\begin{figure*}[t]
    \centering
    \subfloat[Effect of horizon length and replan interval\label{fig:horizon_tradeoff}]{
        \includegraphics[height=4.4cm]{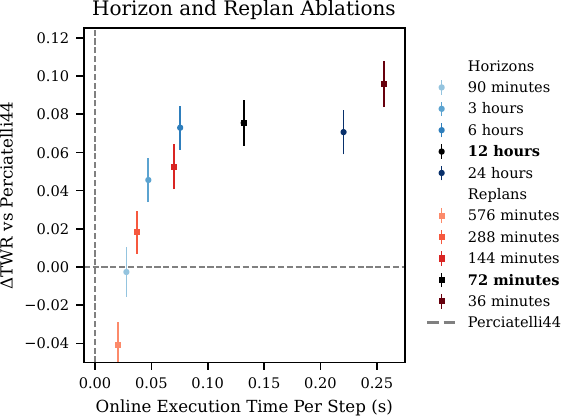}}
    \subfloat[Effect of number of initializations\label{fig:init_tradeoff}]{
        \includegraphics[height=4.4cm]{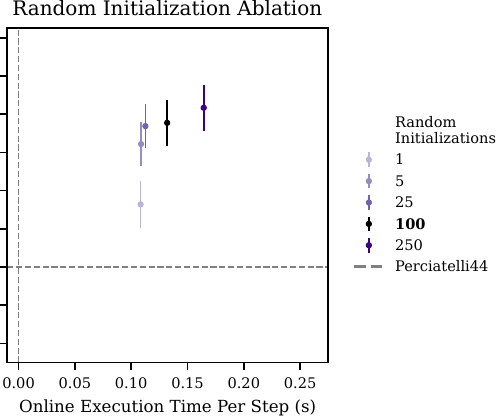}}
    \subfloat[Effect of dynamics fidelity and wind model\label{fig:fidelity_tradeoff}]{
        \includegraphics[height=4.4cm]{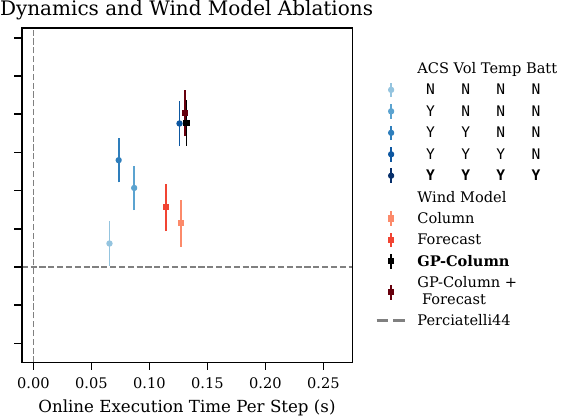}}
    \caption{Performance–efficiency trade-offs in FOMPC. Each plot reports mean time-within-radius relative to Perciatelli44 (error bars show 95\% confidence intervals) as a function of online execution time per step. Markers denote parameter settings; bolded values indicate defaults (12h horizon, 72m replan interval, 100 initializations, highest dynamics fidelity, GP-Column wind model). Each plotted setting varies one parameter at a time from this default.
    }
    \label{fig:mpc_ablation_tradeoffs}
\end{figure*}

We explore how FOMPC performance compares to Perciatelli44 under identical conditions and how the hyperparameters of FOMPC affect its own performance and computational efficiency. Perciatelli44 is the implementation of the state-of-the-art QR-DQN station-keeping controller described in \cite{bellemare2020autonomous, agarwal2022reincarnating,dabney2018distributional} and implemented in the Balloon Learning Environment (BLE) \cite{Greaves_Balloon_Learning_Environment_2021}. 

After running FOMPC and Perciatelli44 in BLE with the same set of conditions, we measure performance metrics such as time-within-radius (TWR). We also vary the control parameters and modeling assumptions of FOMPC and evaluate under these conditions to understand how they affect FOMPC performance.
\vspace{-0.25pt}

\subsection{Experimental Setup}
\label{subsec:experimental_setup}
We begin by describing the evaluation protocol. All experiments are conducted in BLE on the station-keeping task. Each trial consists of 960 steps, with $\Delta t=$ 3 minutes of simulation time between steps, totaling 2 days of simulated flight time. Additionally, each trial is initialized with a seed, which generates a realistic balloon starting state and unique wind conditions. The time and location of the trial are randomly selected, also based on the seed, uniformly between 2011-1-1 and 2014-12-31 and between $[-10, 10]$ latitude and $[-175, 175]$ longitude. The randomness is seeded from the seed of the whole episode. After every step and at the start of each episode, an agent is given a set of features about the balloon and environment and produces an action. Each agent we evaluate is measured on the same set of randomized wind field seeds to ensure comparability. After running each seed, we collect the TWR \eqref{problem_def}, online execution time per step (i.e., time it takes an agent to calculate an action), power safety layer violations (Sec.~\ref{sec:battery}), and the Perciatelli44 reward over all the steps. The Perciatelli44 reward function counts the steps within the radius and has an exponential drop-off term for steps outside the radius, with an additional power regularization term for safe power usage \cite{Greaves_Balloon_Learning_Environment_2021}.

For FOMPC parameter ablations, each agent is evaluated on seeds $[0, 1000)$ in BLE. This range provides a broad distribution of conditions that captures the variability of real-world stratospheric winds, from featureless flows that are difficult to navigate to diverse winds that lend themselves to station-keeping. We collected a larger set of seeds $[0, 10\,000)$ for Perciatelli44 and our default FOMPC to provide a more thorough comparison. For all results, we report the mean of these metrics across trials and the $95\%$ confidence interval computed by $\pm 1.96(\frac{\text{variance}}{\text{number of trials}})^{1/2}$. Seed sets $[0, 1000)$ and $[0, 10\,000)$ are provided as evaluation benchmarks in BLE \cite{Greaves_Balloon_Learning_Environment_2021}.

Each agent receives features from the environment before computing an action. The feature set for FOMPC is the current state $\mathbf {x}_{t_n}$ and wind model $\hat{W}^\psi$, which may be initialized with state $\mathbf{x}_{t_n}$. The feature set for Perciatelli44 is the Gaussian process (GP)-refined wind column, $\hat{W}^{\psi_2}_\mathbf{x_{t_n}}$ evaluated at a pre-defined set of reachable pressure levels. This set of wind values is concatenated with the balloon state $\mathbf{x}_{t_n}$ in one feature vector.

\subsection{Overall Performance (FOMPC vs Perciatelli44)}
\label{subsec:overall_performance}

We define a default FOMPC by selecting hyperparameters with high performance and reasonable computational cost. These are a 12-hour planning horizon ($H= 240$ steps), 72-minute replanning interval ($m = 24$ steps), 100 random initializations, $\phi_4$ dynamics fidelity level, and $\psi_2$ wind model, the GP-refined wind column.

To test whether a model-based planner can achieve station-keeping performance competitive with RL policies, we compare default FOMPC to Perciatelli44.

\begin{table}[h]
\centering
\begin{tabular}{lcc}
\hline
\textbf{Metric} & \textbf{FOMPC} & \textbf{Perciatelli44} \\
\hline
Reward & $513.27 \pm 5.36$ & $425.89 \pm 5.25$ \\
Time Within Radius (\%) & $39.43 \pm 0.65$ & $31.78 \pm 0.60$ \\
Execution Time per Step (s) & $0.131 \pm 0.000$ & $0.00110 \pm 0.000$ \\
Power Safety Violations & $8.70 \pm 0.55$ & $22.14 \pm 0.84$ \\
\hline
\end{tabular}
\caption{Comparison between FOMPC and Perciatelli44.
Values are reported as mean $\pm$ 95\% confidence interval over 10k trials.}
\label{tab:baseline_comparison}
\end{table}

Table~\ref{tab:baseline_comparison} shows that our FOMPC consistently outperforms Perciatelli44 in terms of RL reward and TWR, while also reducing power safety violations. This improvement comes at the expense of increase in computation time per control step. However, this time per step is still much less than the replan interval of 72 minutes between when new actions are needed. 

\subsection{Ablation Studies}
Next, we investigate how individual FOMPC hyperparameters influence TWR performance and execution time per step. In each ablation, we vary a single parameter while holding all other parameters at the default setting. This controlled setup isolates the effect of each design choice. We consider five factors: planning horizon, replan interval, number of initializations, dynamics fidelity, and wind model.

\subsubsection{Planning Horizon}

We vary the planning horizon across five values: 90 minutes, 3 hours, 6 hours, 12 hours (default), and 24 hours ($H= 30, 60, 120, 240, 480$). Longer horizons allow the controller to anticipate wind patterns further into the future, but also increase computational cost. As shown in Fig.~\ref{fig:horizon_tradeoff}, TWR improves steadily up to a horizon of 12 hours, but drops off when extended to 24 hours, indicating that overly long horizons can degrade performance despite higher computation. All but the shortest horizon outperform Perciatelli44.

\subsubsection{Replan Interval}

We vary the replan interval across five values: 576 minutes, 288 minutes, 144 minutes, 72 minutes (default), and 36 minutes ($m= 192, 96,48, 24, 12$). Shorter replan intervals allow the controller to adapt more frequently to changing wind conditions, but also increase computational cost. As shown in Fig.~\ref{fig:horizon_tradeoff}, performance improves as the replan interval decreases, but gains reflect diminishing returns at higher computational cost.
All but the longest replan outperform Perciatelli44.

\subsubsection{Number of Initializations}

We vary the number of random optimization initializations across five settings: 1, 5, 25, 100 (default), 250. Considering more initializations improves the chances of avoiding poor local minima in the optimization process but increases runtime. Fig.~\ref{fig:init_tradeoff} shows a performance improvement with diminishing returns at higher computational costs. Even a single random initialization outperforms Perciatelli44.

\subsubsection{Dynamics Fidelity}

We vary the level of dynamics fidelity across five settings: $\phi_0, \phi_1, \phi_2, \phi_3, \phi_4$ (default; see Table~\ref{tab:fidelity}). Lower-fidelity dynamics reduce computational cost by simplifying the balloon dynamics, but may degrade performance if important physical effects are omitted. As shown in Fig.~\ref{fig:fidelity_tradeoff}, TWR improves as fidelity increases, but there are diminishing returns. Even the simplest fidelity level outperforms Perciatelli44.

\subsubsection{Wind Model}

We vary the wind model across four models: Forecast, Column, GP-Column (default), GP-Column + Forecast ($\psi = \psi_0, \psi_1,\psi_2 \text{ (default)}, \psi_3$, see Sec.~\ref{subsec:wind_model}). Fig.~\ref{fig:fidelity_tradeoff} reveals the importance of online estimation of forecast errors.
FOMPC with the GP-informed wind column $\hat{W}^{\psi_2}$ outperforms that with the full forecast $\hat{W}^{\psi_0}$ despite the latter's richer spatiotemporal information,
while the combined model $\hat{W}^{\psi_3}$ (marginally) outperforms both.
This benefit comes at the cost of additional online planning time to fit the Gaussian Process.
We also note that the timing difference between column and forecast models is negligible due to their similar implementations in JAX. All wind models demonstrate improvement over Perciatelli44.

\subsection{Discussion}

Our experiments highlight that FOMPC, a model-based controller, outperforms the state-of-the-art RL controller on TWR while more closely adhering to power safety constraints. Many hyperparameter configurations still surpass Perciatelli44, with performance degrading only under extremely constrained horizons or replan intervals. Even lightweight dynamics and limited wind information retain their performance. These gains come at the cost of higher per-step computation, but this is practical in the balloon domain, where actions are required only every few minutes rather than at millisecond frequencies typical in other MPC applications. It is also the case that many balloons are controlled remotely with action sequences computed elsewhere and sent to the balloon via satellite communication. In addition, unlike Perciatelli44, FOMPC requires no training data and incurs no upfront training cost. However, further experimentation (Fig.~\ref{fig:app_robustness}) shows Perciatelli44 is more robust to unreliable forecasts; conversely, FOMPC overfits and degrades under high-wind noise.

The ablations show that incorporating learned forecast corrections into the wind model yields the largest gains in TWR. Interestingly, spatiotemporal forecast data alone offered less benefit for long-horizon planning, suggesting that compensating for forecast bias is more impactful than merely extending wind information.

To better understand RL's failure modes, we first compare performance under the Perciatelli44 reward. As shown in Table~\ref{tab:baseline_comparison}, FOMPC achieves higher scores than Perciatelli44 even on this reward, despite optimizing a different surrogate for TWR. This indicates that the reward used by Perciatelli44 is a reasonable proxy for TWR, and its lower performance cannot be attributed to an objective mismatch. We also note that $\hat{W}^{\psi_2}$ contains the same information as Perciatelli44, yet our model-based controller outperforms Perciatelli44, meaning the measurement-informed wind column is sufficient for high-performing TWR. Two remaining hypotheses for the Perciatelli44 performance are 1) a limited policy class or 2) under-optimization. Our experiments do not let us form a conclusion on whether the limitations stem primarily from the expressiveness of the Q-network architecture or from inadequate training.

\section{CONCLUSION}
In this work, we revisited the problem of high-altitude balloon station-keeping through the lens of MPC, due to its absence in the literature as a baseline for RL and its potential to generalize across new dynamics and unseen wind conditions. By formulating the balloon dynamics as differentiable functions in JAX and applying FOMPC, we established a model-based controller that consistently outperforms the state-of-the-art RL controller Perciatelli44 in the BLE. Our experiments demonstrate that planning over a horizon, even when restricted to the same wind information, leads to more effective control decisions, yielding significant gains in TWR while also reducing power safety violations.

Through systematic ablations of planning horizon, replan interval, initialization strategy, dynamics fidelity, and wind models, we identified which factors contribute most to performance and where computational efficiency can be traded against accuracy. These insights highlight that, given the same wind information, MPC can outperform Perciatelli44 on both its own reward and time within radius. It is possible that Perciatelli44's QR-DQN structure is not able to represent long-horizon strategies or adapt to wind conditions in the way FOMPC can. Alternatively, its performance may be limited by under-optimization, as training may not fully converge given the complexity of the environment.

To investigate, future work can use MPC as a teacher for the Perciatelli44 policy network. If imitation learning enables the policy to match MPC performance, this would indicate that the network class can represent long-horizon solutions but was under-optimized. If not, it would suggest that this type of Q-network is insufficiently expressive for the task. More broadly, this points to opportunities for hybrid approaches where reinforcement learning can distill or imitate MPC, combining the generalization ability and runtime efficiency of policies with the foresight of model-based planning. 

There are additional opportunities for future work on pure MPC approaches, such as addressing robustness to uncertainty and runtime.
In additional experiments presented in the appendix, we observe that FOMPC remains sensitive to extreme noise (Fig.~\ref{fig:app_robustness}). Model Predictive Path Integral (MPPI) \cite{williams2015mppi} is a scalable alternative which can address computation time via hardware parallelization, but it currently fails to stay in safe regions (Table~\ref{tab:mppi_mpc_comparison}).

Our differentiable balloon model and strong MPC baselines thus provide a foundation for advancing navigation in the stratosphere  and other underactuated flow-driven systems.

\section{ACKNOWLEDGMENTS}
This work was partly funded by NSF CCF 2312774, NSF OAC-2311521, NSF IIS-2442137, a gift to the LinkedIn-Cornell Bowers CIS Strategic Partnership, the AI2050 Early Career Fellowship program at Schmidt Sciences, and a 2030 Project Fast Grant from the Cornell Atkinson Center for Sustainability.



\vspace{-4pt}
\appendix

\subsection{Robustness to Wind Uncertainty}

\begin{figure}[H]
    \centering
    \includegraphics[width=1.0\columnwidth]{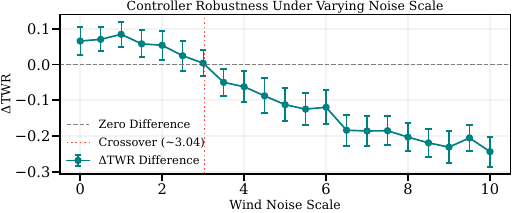}
    \caption{TWR sensitivity of FOMPC and Perciatelli44 to scaled wind magnitude.}
    \label{fig:app_robustness}
    \vspace{-10pt}
\end{figure}

We evaluate the robustness of FOMPC and Perciatelli44 to varying levels of unmodeled disturbances in the real wind. We define new wind fields $W = W_0 + \mu \cdot b$ (Eq.~\ref{eq:wind_field_def}) by varying $\mu$, where $\mu$ is the noise scale and $W_0$ is the forecast. For each $\mu$, we evaluate on seeds $[0, 100)$. Fig.~\ref{fig:app_robustness} shows the time-within-radius difference between FOMPC and Perciatelli44 ($\Delta\text{TWR}$) and $95\%$ confidence intervals. FOMPC performance degrades relative to Perciatelli44, with a crossover occurring at $\mu \approx 3$, triple the typical noise level.
 
We note that changing the underlying wind field $W$ changes the nature of a trial. Therefore, across different values of $\mu$, a set of seeds is no longer comparable, so we only consider the difference in performance of FOMPC and Perciatelli44 for a specific $\mu$ value. 

Performance of FOMPC degrading under wind noise is consistent with limitations of deterministic planning. As in Eq.~\ref{eq:mpc}, the cost is determined by state trajectory of the balloon under the wind model. This cost is irrespective of the quality of the wind model or the uncertainty of specific wind measurements that determined the state trajectory. Thus, FOMPC trusts the wind measurements, and thereby the cost of a trajectory, for every $\mu$ level. Future work may consider using uncertainty quantification from the Gaussian Process (used to predict forecast errors) in the MPC formulation (Eq.~\ref{eq:mpc}).

Conversely, Perciatelli44's performance is relatively robust to wind at higher noise scales. This potentially may be attributed to its state representation, as wind magnitudes are squashed to a maximum value before used as a network input. This may act as a regularizer, preventing extreme wind vectors from pushing the network into out-of-distribution states. It is unclear if Perciatelli44's offline training included winds of this magnitude, but the squashed feature space, in theory, limits extrapolation errors. This may allow the policy to fall back on generalized station-keeping behaviors it learned in training.

\subsection{MPPI Performance and Limitations}

We also test MPPI \cite{williams2015mppi} on seeds $[0, 1000)$ and report the time-within radius and 95\% confidence intervals. MPPI performs competitively to Perciatelli44 but is susceptible to altitude-related early termination (over-pressurization). MPPI terminates early in 92 seeds (Perciatelli44 and FOMPC do not terminate early) and its time-within-radius is reported as if the balloon were outside the radius for the remaining steps of the episode after termination.

\begin{table}[H]
\centering
\small
\setlength{\tabcolsep}{4pt}

\captionsetup{
    font=small,
    labelfont=bf,
    skip=6pt
}

\begin{tabular}{lccc}
\hline
\textbf{Metric} & \textbf{FOMPC} & \textbf{Perciatelli44} & \textbf{MPPI} \\
\hline
TWR (\%) & $37.10 \pm 2.02$ & $29.55 \pm 1.84$ & $30.11 \pm 1.86$ \\
Time / Step (s) & $0.132 \pm 0.00$ & $0.001 \pm 0.00$ & $0.089 \pm 0.00$ \\\hline
\end{tabular}
\caption{Comparison of FOMPC, Perciatelli44, and MPPI.}
\label{tab:mppi_mpc_comparison}
\vspace{-10pt}
\end{table}

MPPI operates with the same cost function as Eq.~\ref{eq:cost_fn}. Planning is done with 512 samples per computed action. Actions are sampled at 24 evenly spaced points over a 240-step planning horizon with a standard deviation of 0.3 interpolated with a zero-order hold. The initial plan of each trial is all 0. 

Since the difficulty of each trial depends on the wind, we use a binary search over the range $[0.001, 10000.0]$ to find an appropriate temperature value $\lambda$ on each replan. We search for a $\lambda$ such that the approximate effective sample size $\frac{1}{512 \cdot \sum_{i=1}^{512} w_i^2}$ (where $w_i = \frac{\exp(-C_i / \lambda)}{\sum_{j=1}^{512} \exp(-C_j / \lambda)}$ and $C_i$ represents the rollout costs) is 0.2.

Further engineering, like adding more safety terms to the cost function, input transformations (as in Sec.~\ref{subsec:cont_action_constriction}), or action noise scheduling may yield further improvements in MPPI performance.
 
\subsection{Review of Finetuned Perciatelli44}

While a \textit{Finetuned Perciatelli44} agent is reported to outperform the standard \textit{Perciatelli44} baseline in several sources, no implementation details or official release are available. The Balloon Learning Environment (BLE) documentation~\cite{ble_docs} includes a training performance plot where \textit{Finetuned Perciatelli44} surpasses \textit{Perciatelli44}, though it is evaluated on only 100 seeds compared to 10{,}000 for \textit{Perciatelli44}. In our experiments, evaluation over 10{,}000 seeds consistently yields higher average performance than smaller (100–1{,}000) seed evaluations, suggesting that \textit{Finetuned Perciatelli44} would likely achieve above 34\% TWR under comparable conditions.  

A second reference appears in the BLE blog post~\cite{ble_blog}, where \textit{Finetuned Perciatelli44} reportedly reaches $\approx$38\% TWR versus $\approx$32\% for \textit{Perciatelli44}. Although the evaluation setup is not described, the $\approx$32\% baseline aligns closely with our own 10{,}000-seed results. Finally, \textit{Reincarnating RL}~\cite{agarwal2022reincarnating} presents a teacher-normalized score (distinct from TWR) evaluated on 10{,}000 seeds with \textit{Perciatelli44} as the teacher. There, \textit{Finetuned Perciatelli44} achieves 107\%, corresponding to roughly $\approx$34\% TWR.  

Assuming \textit{Finetuned Perciatelli44} achieves $\approx$38\% under the 10{,}000-seed evaluation, our best agent configuration—identified through ablations—reaches $\approx$39\% on the 1{,}000-seed evaluation and may perform even better at 10{,}000 seeds (as observed for both the default FOMPC parameters and \textit{Perciatelli44}). This places our agent approximately 5–10\% above the highest-reported \textit{Finetuned Perciatelli44} performance, though further confirmation would require access to the missing implementation details.

\end{document}